\documentclass[11pt]{article}

\usepackage{amsmath,amsthm,verbatim,amssymb,amsfonts,amscd, graphicx}
\usepackage{graphics}
\usepackage{authblk}
\usepackage{bm}
\usepackage{algorithmic}
\usepackage[lined,boxed]{algorithm2e}
\usepackage{hyperref}

\usepackage{wrapfig}
\usepackage{subfigure}

\theoremstyle{plain}
\newtheorem{theorem}{Theorem}

\newtheorem{proposition}{Proposition}

\theoremstyle{definition}

\begin{document}

\title{$k$-means: Fighting against Degeneracy in Sequential Monte Carlo with an Application to Tracking}
\author[1]{
\textbf{Kai Fan} 
}
\author[1]{
\textbf{Katherine Heller} 
}
\affil[1]{Duke University}
\affil[1]{\texttt{\{kai.fan,kheller\}@stat.duke.edu}}

\maketitle

\begin{abstract}
\begin{quote}
For regular particle filter algorithm or Sequential Monte Carlo (SMC) methods, the initial weights are traditionally dependent on the proposed distribution, the posterior distribution at the current timestamp in the sampled sequence, and the target is the posterior distribution of the previous timestamp. This is technically correct, but leads to algorithms which usually have practical issues with degeneracy, where all particles eventually collapse onto a single particle. In this paper, we propose and evaluate using $k$ means clustering to attack and even take advantage of this degeneracy. Specifically, we propose a Stochastic SMC algorithm which initializes the set of $k$ means, providing the initial centers chosen from the collapsed particles. To fight against degeneracy, we adjust the regular SMC weights, mediated by cluster proportions, and then correct them to retain the same expectation as before. We experimentally demonstrate that our approach has better performance than vanilla algorithms.
\end{quote}
\end{abstract}

\section{Introduction}

$K$-means clustering is a special case of Gaussian Mixture Models (GMMs) assuming fixed covariance matrix $\sigma^2\mathbf{I}$ ($\mathbf{I}$ is the identity matrix), thus requiring the only estimator to be cluster centers $\mu_{1:k}$. Referred to as the vanilla "$k$-means", the algorithm basically begins with randomly chosen $k$ points from the dataset. Using these points as initial centroids, it then assigns each data point to have membership in the nearest centroid (usually based on Euclidean Measurement), and next computes new centroids by averaging the data points in each cluster together. The final assignments are made when each data point's membership is unchanged. This is called hard Expectation Maximization (EM). However, this strategy of random initialization leads to local minima. To relieve this problem, $k$-means++ \cite{arthur2007k} introduced a sequentially random selection strategy according to a squared distance from the closest center already selected. This trick can theoretically guarantee that the expectation of distortion is no more than $\mathcal{O}(\log k)$ worse than optimal. \cite{figueiredo2002unsupervised} also proposed a heuristic approach with incrementally growing GMMs by splitting large clusters.

In our work, taking advantage of the degeneracy of SMC, we also propose a seemingly paradoxical algorithm to make the initialization of $k$-means more robust. Intuitively, estimating the centers is equivalent to learning the means of Gaussian Mixture models. However, as in initialization, it is not essential to provide a set of means exactly to the model. It is sufficient for initial centroids to fall into potential clusters, since a local minima is likely to be reached if at least two initial centroids are close in distance. This can be achieved by leveraging the quick degeneracy of SMC.   

SMC methods \cite{kitagawa2012smoothness} are a class of simulation-based techniques which use importance sampling and resampling mechanisms to compute posterior distributions and do parameter estimation, typically for a state-space model. In a regular clustering task, data points do not sequentially stream in, but some offline SMC methods \cite{fernandez2007estimating,andrieu2010particle,ionides2006inference} were designed to plug into a standard Metropolis-Hastings (MH) Sampler for approximating intractable distributions. We adopt a Bayesian formulation for parameter estimation as well, but discard the MCMC sampling and full dataset training, both of which are time consuming. In contrast, we follow a popular convention --  stochastic training, starting with randomly dividing the dataset into a number of batches, and feeding them sequentially to a standard SMC algorithm. Such attempts have been widely implemented in optimization, such as in stochastic gradient descent \cite{bubeck2014theory} or stochastic variational inference \cite{hoffman2013stochastic}. Our proposed SSMC is also designed to benefit from the computational efficiency of stochastic approach.

On the other hand, the more reasonable way to intuit using the degeneracy property to initialize is that the particles of SMC methods usually have non-negligible weights if they are around the modes of posterior distribution. In practice, the appearance of this phenomenon can be mitigated, but is difficult to completely solve. From this viewpoint, we are interested in relieving degeneracy via finding the largest possible number of modes in the posterior distribution. Addressing this issue has been intensively studied in recent works. \cite{gilks2001following} proposed to introduce diversity among particles by moving the particles using MCMC moves with an invariant distribution. \cite{doucet2006efficient} suggested a block sampling strategy to reduce the number of resampling steps and thus, the possibility of degeneracy. \cite{lin2013lookahead} provided a comprehensive review for various lookahead strategies used in fighting degeneracy. \cite{yang2013sequential} refined SMC to a Sequential Markov Chain Monte Carlo to proceed as in usual MCMC but with the stationary distribution updated appropriately each time a new instance streams in.

However, the intuition above can contribute to another possible solution to mitigate degeneracy. Clustering the particles can force them to have own memberships, thus inducing a set of new weights with respect to cluster partitions. We can balance the original SMC weights with new cluster weights for the resampling step, and then correct them to keep the same expectation. In addition, the geometric structure of particle filters cannot easily be depicted by the weighted empirical distribution alone, especially in high dimensions. Some potential modes may exist in the subspace of the filters. Thus our approach can extend to a subspace clustering based techniques to adjust the SMC. In particular, we won't specify the clustering method in this modified SMC, since any clustering method can be applied. In fact, we do not care whether the cluster result is globally optimal or not, the only concern is that the less weighted particles have a larger probability of being sampled, under the constraint of an unbiased expectation.  

\section{Preliminary}
\label{sec:pre}

We briefly introduce the formulation of Sequential Monte Carlo, including the bootstrap filtering implementation under a Sequential Importance Resampling (SIR) scheme, laying the foundation for our modification.

The general setting of SIR contains time-series or sequential observations $\mathbf{y}_t \in \mathcal{Y}^e, t=1,...,T$ and hidden states $\mathbf{x}_t \in \mathcal{X}^d,t=0,...,T$. They can be modeled as a Markov Process with an initial distribution $\pi(\mathbf{x}_0)$ and two transition kernels, $\pi(\mathbf{x}_t|\mathbf{x}_{t-1})$ and $\pi(\mathbf{y}_t|\mathbf{x}_t)$. Inference recursively estimates the posterior distribution $p(\mathbf{x}_{0:t}|\mathbf{y}_{1:t})$ and its marginal distribution $p(\mathbf{x}_t|\mathbf{y}_{1:t})$. The expectation $\mathbb{E}_{p(\mathbf{x}_{0:t}|\mathbf{y}_{1:t})}f_t(\mathbf{x}_{0:t})$ is computed, where $f_t:\mathcal{X}\rightarrow\mathbb{R}^{n_{f_t}}$ can be any integrable function. In the case of a tracking problem \cite{gustafsson2010particle}, setting a special function $f_t(\mathbf{x}_{0:t})=\mathbf{x}_{0:t}$, filtering can find an object path by computing the mean of potential location $\mathbf{x}_{0:t}$. An experimental tracking example will discussed in a later section.

\subsection{Sequential Importance Resampling}

\begin{algorithm}[t]
\label{alg:BF}
\caption{Bootstrap Filter}
\begin{algorithmic}[1]
\STATE $t=0$, Sample $X_0^{(i)} \sim p(\mathbf{x}_0)$, $i=1,\ldots,N$.
\WHILE{$t\geq1$}
\FOR{each $i$}
\STATE Sample $\tilde{X}_t^{(i)} \sim \pi(\mathbf{x}_t|\mathbf{x}_{t-1}=X_{t-1}^{(i)})$;
\STATE $w_t^{(i)} \propto w_{t-1}^{(i)}\frac{\pi(\mathbf{y}_t|\mathbf{x}_t=\tilde{X}_t^{(i)})\pi(\mathbf{x}_t=\tilde{X}_t^{(i)}|\mathbf{x}_{t-1}=X_{t-1}^{(i)})}{\pi(\mathbf{x}_t=\tilde{X}_t^{i}|\mathbf{y}_{1:t},\mathbf{x}_{0:t-1}=X_{0:t-1}^{i})}$;
\ENDFOR
\STATE Normalize $\{w_t^{(i)}\}_{i=1}^N$;
\STATE $\hat{p}(\mathbf{x}_{0:t}|\mathbf{y}_{1:t})=\sum_{i=1}^Nw_t^{(i)}\delta_{\tilde{X}_{0:t}^{(i)}}(\mathbf{x}_{0:t})$, where $\tilde{X}_{0:t}^{(i)}=\{X_{0:t-1}^{(i)},\tilde{X}_t^{(i)}\}_{i=1}^N$;
\IF{ $\hat{S}_{ess}=\left(\sum_{i=1}^N(w_t^{(i)})^2\right)^{-1} < \frac{N}{2}$} 
\STATE Resample $X_{0:t}^{(i)} \sim \hat{p}(\mathbf{x}_{0:t}|\mathbf{y}_{1:t})$ \STATE Update $\hat{p}(\mathbf{x}_{0:t}|\mathbf{y}_{1:t})=\frac{1}{N}\sum_{i=1}^N\delta_{X_{0:t}^{(i)}}(\mathbf{x}_{0:t})$.
\ENDIF
\STATE $t \leftarrow t+1$;
\ENDWHILE
\end{algorithmic}
\end{algorithm}

Regular importance sampling allows the integral $\mathbb{E}_{p(\mathbf{x}_{0:t}|\mathbf{y}_{1:t})}f_t(\mathbf{x}_{0:t})$ to be constructed with weight function 
$
w(\mathbf{x}_{0:t})=\frac{p(\mathbf{x}_{0:t}|\mathbf{y}_{1:t})}{\pi(\mathbf{x}_{0:t}|\mathbf{y}_{1:t})} $, 
where $\pi(\mathbf{x}_{0:t}|\mathbf{y}_{1:t})$ is the proposal distribution. However, the Monte Carlo estimate $\sum_{i=1}^Nf_t(\mathbf{x}_{0:t}^{(i)})\tilde{w}_t^{i}$ requires the simulated particles $\{\mathbf{x}_{0:t}^{(i)}\}_{i=1}^N$ in accordance with $p(\mathbf{x}_{0:t}|\mathbf{y}_{1:t})$, where $\tilde{w}_t^{i}$ is normalized $w(\mathbf{x}_{0:t})$. This estimate implicitly leads to a high cost of computation when recomputing the distribution at each iteration for newly arrived observation $\mathbf{y}_{t+1}$. Therefore, sequential importance sampling (SIS) circumvents such a deficiency by admitting the assumption that proposal distribution $\pi(\mathbf{x}_{0:t}|\mathbf{y}_{1:t})$ has a decomposition form including marginal distribution at time $t-1$. In other words, we have
\begin{align}\label{eq:assume}
\pi(\mathbf{x}_{0:t}|\mathbf{y}_{1:t}) = \pi(\mathbf{x}_t|\mathbf{x}_{0:t-1},\mathbf{y}_{1:t}) \pi(\mathbf{x}_{0:t-1}|\mathbf{y}_{1:t-1})
\end{align}
where Equation (\ref{eq:assume}) indicates the importance weights can be simplied to a recursion formulation.
\begin{align}\label{eq:weight}
\tilde{w}_t\propto \tilde{w}_{t-1}\frac{\pi(\mathbf{y}_t|\mathbf{x}_t)\pi(\mathbf{x}_t|\mathbf{x}_{t-1})}{\pi(\mathbf{x}_t|\mathbf{x}_{0:t-1},\mathbf{y}_{1:t})}
\end{align}
Hence, if one can simulate $N$ particles according to the proposal distribution, we can interpret the sampling method as an alternative approximation of posterior \begin{align}
p(\mathbf{x}_{0:t}|\mathbf{y}_{1:t})\approx \sum_{i=1}^N \tilde{w}_t^{(i)}\delta_{X_{0:t}^{(i)}}(\mathbf{x}_{0:t})
\end{align}
where $\delta_{\cdot}(\cdot)$ refers to the Dirac mass function. This estimate is biased for a fixed $N$ but asymptotically consistent. The derived bootstrap filter (BF) algorithm can be seen in Algorithm \ref{alg:BF} or \cite{doucet2001introduction}.

\subsection{Degeneracy}

In a standard BF algorithm, \cite{doucet2006efficient} indicated that the variance of importance weights  becomes large over several iterations, which refers to a quantitative measurement of degeneracy. The sequential importance sampling algorithm may fail after a few steps because the marginal distribution collapses, that is to say, all but a few particles will have negligible importance weights. Concretely, degeneracy means for any $N$ and any $t$, there exists $n(t,N)$ such that for any $n>n(t,N)$, $\hat{p}(\mathbf{x}_{0:t}|\mathbf{y}_{1:n})=\delta_{X_{0:t}^*}(\mathbf{x}_{0:t})$, collapsing to a unique point distribution. \cite{liu1998sequential} suggested that $\hat{S}_{ess}=\left(\sum_{i=1}^N(w_t^{(i)})^2\right)^{-1}$ provides an implementable approximation to measure the degeneracy. Apparently, ESS takes a value between 1 and $N$; thus most works suggest that $N/2$ is a reasonable threshold to determine whether the degeneracy happens but it can be tuned depending on the specific application. If the weights have large variance, the distribution of weights will be unbalanced or have small entropy. A resampling operation is designed to choose particles with equal probability to their current weights and assign uniform weights to the new particle set, referred to as Sequential Importance Resampling together with SIS. This scheme does not actually solve the degeneracy problem but reduces the variance of the weights, since the particles with high weights will produce more offspring in the resampling procedure. Sample impoverishment still exists due to eliminating the particles with low weights, with high probability. 

\section{Algorithms}
\label{sec:alg}

In this section, we will first introduce the idea that the disadvantage of SMC can be seen as an advantage in the initialization of the parameters of the Gaussian Mixture Models. Next, we will reverse this process and explore how to adjust the regular weights in standard SMC by leveraging cluster methods. 

\subsection{Stochastic SMC}

\begin{wrapfigure}{r}{0.3\textwidth}
  \vspace{-20pt}
  \subfigure{
    \includegraphics[width=40mm,clip,trim=50 25 50 10mm]{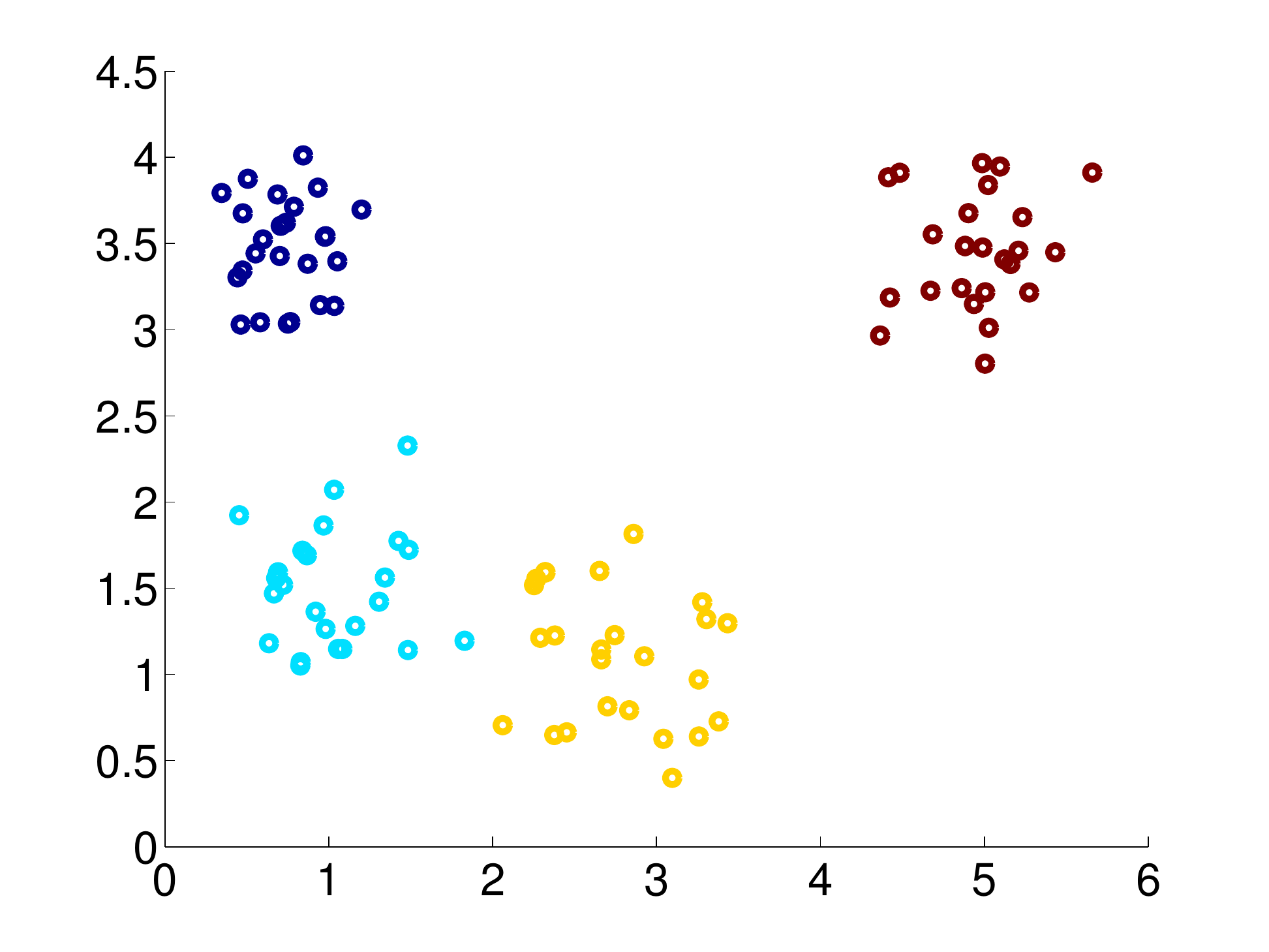}}
    \subfigure{
\includegraphics[width=40mm,clip,trim=50 25 50 10mm]{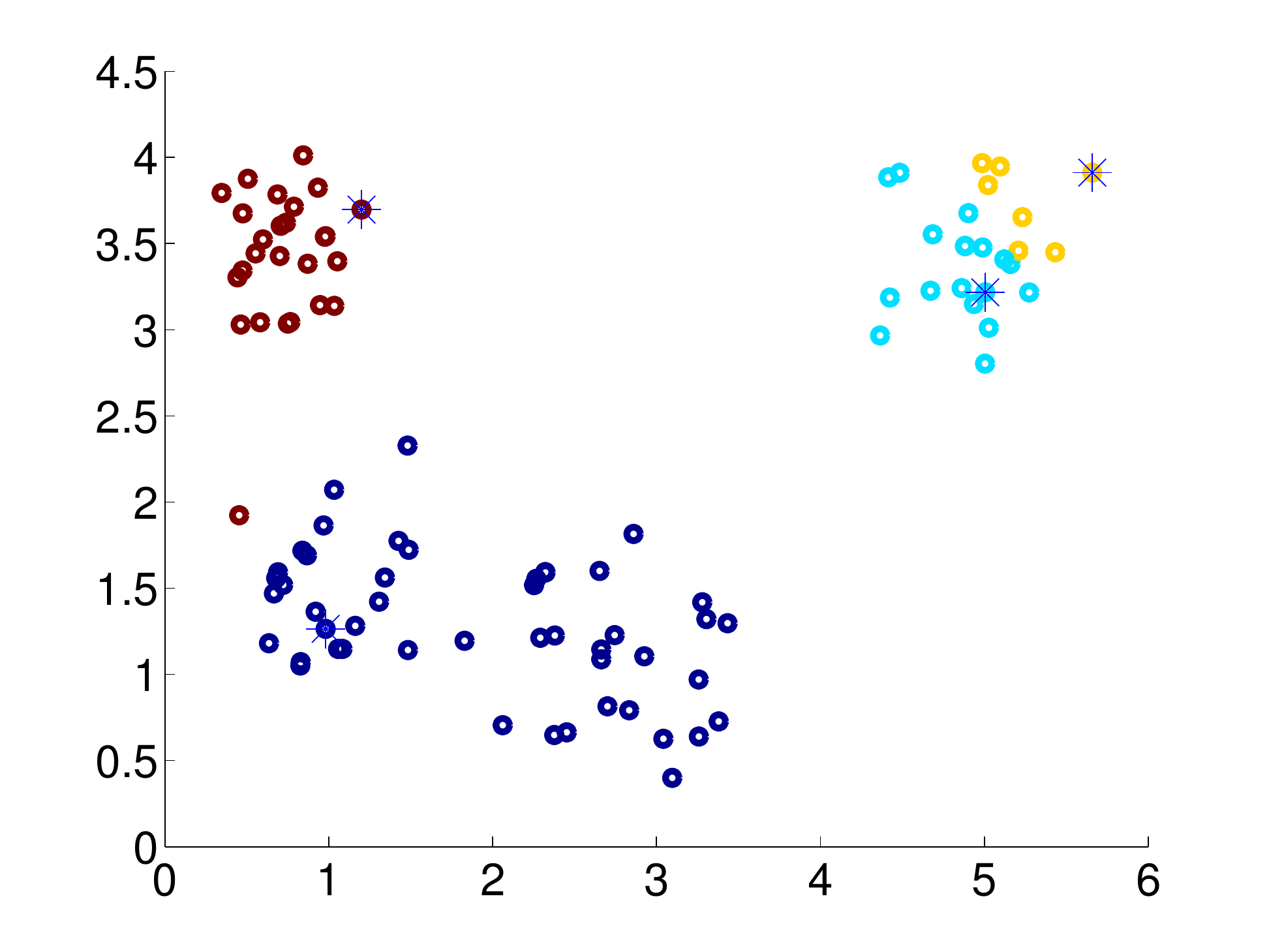}
}
  \vspace{-10pt}
  \caption{An example}
  \vspace{-10pt}
\label{fig:bad}
\end{wrapfigure}

It is often observed that $k$-means starting with a bad initialization will converge to a local minima. Fig.\ref{fig:bad} shows a typical example of this phenomenon, where the top panel represents the true membership of four clusters, and the bottom one shows one convergence result of $k$-means with a random initialization, and the stars are the initial configuration of centroids. Intuitively, this happens when multiple initial centers stay too close, especially within same underlying cluster.

From a Bayesian perspective, the dataset $\{\mathbf{y}_i\}_{i=1}^N$ can be interpreted as sampling from a generative model. The $k$-means setting restricts the generative model of GMMs to share the common covariance matrix $\sigma^2I$, and equal mixing weights $\pi_{1:k}=1/k$, thus requiring the only goal to estimate the mean of each Gaussian distribution, $\mu_{1:k}$. We can then estimate $\Theta$ or $p(\Theta|\mathbf{y}_{1:N})$, where $\Theta$ denotes all unknown parameters.

It is natural to divide the $\{\mathbf{y}_i\}_{i=1}^N$ into a number of small batches with size $B\ll N$. At each timestamp $t$, we can pretend $\mathbf{y}_{B(t-1)+1:Bt}$ stream in. If $B$ is small, the posterior $p(\Theta|\mathbf{y}_{1:Bt})$ is not very different from $p(\Theta|\mathbf{y}_{1:B(t+1)})$. This simulated annealing trick allows us to formulate an SMC scheme for parameter estimation, particularly to construct the sequential importance weights:
\begin{align}\label{eq:psudow}
w_t^{(i)}&\propto w_{t-1}^{(i)}\frac{p(\Theta^{(i)}|\mathbf{y}_{1:Bt})}{p(\Theta^{(i)}|\mathbf{y}_{1:B(t-1)})}\propto w_{t-1}^{(i)} \frac{p(\mathbf{y}_{1:Bt}|\Theta^{(i)})}{p(\mathbf{y}_{1:B(t-1)}|\Theta^{(i)})} \nonumber \\
&\propto w_{t-1}^{(i)}p(\mathbf{y}_{B(t-1)+1:Bt}|\mathbf{y}_{1:B(t-1)},\Theta^{(i)})
\end{align}
Notice that latent variables indicating the underlying assignment is not included in Eq. (\ref{eq:psudow}). This makes sense because of the property of $k$-means that the latent assignment can be deterministically computed given $\mathbf{y}_{1:B(t-1)}$ and $\Theta^{(i)}$. 

\begin{algorithm}[t]
\label{alg:ssmc}
\caption{Stochastic SMC}
\begin{algorithmic}[1]
\STATE $t=0$, generate candidate particles (Sobol sequence) $\{\Theta_0^{(i)}\}_{i=1}^S$ with uniformly weights $w_0^{(i)}=1/S$;
\REPEAT
\STATE Randomly permute the dataset $\{\mathbf{y}_i\}_{i=1}^N$;
\WHILE{$t\geq1$}
\FOR{each $i$}
\STATE $w_t^{(i)} \propto w_{t-1}^{(i)}p(\mathbf{y}_{B(t-1)+1:Bt}|\mathbf{y}_{1:B(t-1)})$; 
\ENDFOR
\STATE Normalize $\{w_t^{(i)}\}_{i=1}^N$;
\STATE Compute approximate posterior 
$\hat{p}(\Theta_{t}|\mathbf{y}_{1:Bt})=\sum_{i=1}^Sw_t^{(i)}\delta_{\Theta_{t-1}^{(i)}}(\Theta_t)$;
\STATE Sample $\Theta_t^{(i)} \sim \hat{p}(\mathbf{x}_{0:t}|\mathbf{y}_{1:t})$, $i=1,...,S$;
\STATE Update posterior
$\hat{p}(\Theta_t|\mathbf{y}_{1:Bt})=\frac{1}{S}\sum_{i=1}^S\delta_{\Theta_t^{(i)}}(\Theta_t)$;
\IF{ Particles collapse to $k$ unique ones} 
\STATE break; \texttt{/*terminate all loops*/}
\ENDIF
\STATE $t \leftarrow t+1$;
\ENDWHILE
\UNTIL{Particles collapse to $k$ unique ones}
\end{algorithmic}
\end{algorithm}

\begin{algorithm}[t]
 \KwData{$\mathbf{y}_{1:N}$}
 \KwResult{$\mu_{1:k}$}
 \textbf{Initialize} $\mu_{1:k}=\text{SSMC}(\mathbf{y}_{1:N})$\;
 \Repeat{$\mu_{1:k}$ Convergence}{
  $\mathbf{z}_i = \arg\min_{j\in[k]}\|\mathbf{y}_i-\mu_j\|^2$ \texttt{/*Assignment*/}\;
  For each $j\in[k]$, $\mu_j=\frac{1}{\sum_{i:\mathbf{z}_i=j}1}\sum_{i:\mathbf{z}_i=j}\mathbf{y}_i$ \texttt{/*Update means*/}\;
 }
 \caption{$k$-means}\label{alg:skmeans}
\end{algorithm}

However, for general GMMs that contain latent variables $\mathbf{z}$, we may need to construct the joint posterior $p(\Theta,\mathbf{z}_{1:Bt}|\mathbf{y}_{1:Bt})$ by plugging a Metropolis Hastings (MH) algorithm into SMC. However, in $k$-means, we can readily formulate our Stochastic SMC as Algorithm \ref{alg:ssmc}. (see the more details in the supplementary material). Compared with the BF algorithm, SSMC needs only one sampling step (in terms of the resampling step in bootstrap filtering). Since the static parameter is not actually sequential, no transition kernel is introduced such that one step trajectory sampling does not exist before the weights update. Thus Line 9 of Algorithm \ref{alg:ssmc} is weighted particles at the previous time. Additionally, observed data $\mathbf{y}$ pretends to be sequential batches, thus allowing a sequential importance sampling scheme to be constructed. This trick implies a sequence of posterior distributions are generated to propagate the particles. Similar to other stochastic methods, the whole dataset may be reused or not fully used when the termination criterion is met. The third difference is that we replace the degeneracy evaluation criteria. In contrary to other SMC methodology intending to relieve the degeneracy, we expect that the degeneracy will help us to find $k$ initial means. Consequently, the modified $k$-means algorithm with SSMC initialization is stated in Algorithm \ref{alg:skmeans}. Our design is guaranteed  by the theorem proved in \cite{chopin2002sequential}.
\begin{theorem}\cite{chopin2002sequential}
Under mild conditions, if $p(\mathbf{y}_{n+1:n+B}|\mathbf{y}_{1:n},\Theta)$ is bounded, then for any twice differentiable function $f:\Theta\rightarrow\mathbb{R}^{n_f}$, the integrals $\int f(\Theta)\frac{p(\Theta|\mathbf{y}_{1:n+B})^2}{p(\Theta|\mathbf{y}_{1:n})}\mathrm{d}\Theta$ and $\int f(\Theta)f'(\Theta)\frac{p(\Theta|\mathbf{y}_{1:n+B})^2}{p(\Theta|\mathbf{y}_{1:n})}\mathrm{d}\Theta$ both exist, and under the univariate setting, $S/S_{eff}=\mathcal{O}(1)$, if $n\rightarrow\infty$, and $B/n\rightarrow r>0$.
\end{theorem}

To help computational efficiency, candidate particles can be directly set from the dataset. The resultant posterior is required neither to be unbiased nor consistent. Rather than preventing degeneracy, more practical tricks can be developed in application. Though a new method fighting arbitrary initialization, this algorithm also suffers the pitfall that any standard $k$-means approach has -- it fails in high dimensions with significant probability. Some works \cite{timmerman2013subspace,jing2007entropy} attempted to modify the clustering mechanism of $k$-means aiming to obtain subspace results, which SSMC is not capable of; however any existing subspace clustering method can fight the degeneracy.

\subsection{Clustering based SMC}

Recalling the setting of SIR, the degeneracy problem usually arises in practical implementation, and the resampling step induces sample impoverishment. However, if we enlarge the sampling odds of low weighted particles but keep the estimate of the posterior allowing for an invariant expectation, then the degeneracy is, to some extent, postponed for a while. Basically, using any clustering method on a trajectory at time $t$, we obtain the \textit{cumulative weight} $v_t^{(j)}$ of each cluster.
\begin{align}\label{eq:cw}
v_t^{(j)}=\sum_{i:\tilde{X}_t^{(i)}\in C_j}w_t^{(i)}, \forall j\in[k]
\end{align}
where $C_j$ is the $j$th cluster. In practice, the number of clusters $k$ can also be indexed over time $t$.  

Next we follow a parallel sub-resampling step in each cluster according to the normalized weights within cluster. In other words, the weights used for sub-resampling within $j$th cluster are $\frac{w_t^{(i)}\mathbb{I}\{\tilde{X}_t^{(i)}\in C_j\}}{v_t^{(j)}}$, where $\mathbb{I}\{\cdot\}$ is indicator function. Thus the sampling scheme is depicted as:
\begin{align}\label{eq:subsample}
{X'}_{0:t}^{(m)}\sim \sum_{i:\tilde{X}_t^{(i)}\in C_j}\frac{w_t^{(i)}}{v_t^{(j)}}\delta_{\tilde{X}_{0:t}^{(i)}}(\mathbf{x}_{0:t}),m=1,...,|C_j|
\end{align}
This step increases the odds of having particles in all areas of relevant probability without too much of a loss in computation. Even $v_t^{(j)}$ or the weights of some particles are negligible, the relative weights within cluster will become significant after normalization. However, the overall posterior estimation biases the original expectation.

\begin{algorithm}[t]
\label{alg:waa}
\caption{Weight Adjust Algorithm}
\textbf{Input}: $\{X_{0:t-1}^{(i)},\tilde{X}_t^{(i)}\}_{i=1}^N$.\\
\textbf{Output}: $\{w_t^{(i)}\}_{i=1}^N,\{X_{0:t}^{(i)}\}_{i=1}^N,\hat{p}(\mathbf{x}_{0:t}|\mathbf{y}_{1:t})$.\\
\textbf{Algorithm}:
\begin{algorithmic}[1]
\FOR{each $i$}
	\STATE $w_t^{(i)} \propto w_{t-1}^{(i)}\frac{\pi(\mathbf{y}_t|\mathbf{x}_t=\tilde{X}_t^{(i)})\pi(\mathbf{x}_t=\tilde{X}_t^{(i)}|\mathbf{x}_{t-1}=X_{t-1}^{(i)})}{\pi(\mathbf{x}_t=\tilde{X}_t^{i}|\mathbf{y}_{1:t},\mathbf{x}_{0:t-1}=X_{0:t-1}^{i})}$;
\ENDFOR
\STATE Normalize $\{w_t^{(i)}\}_{i=1}^N$;
\STATE $\hat{p}(\mathbf{x}_{0:t}|\mathbf{y}_{1:t})=\sum_{i=1}^Nw_t^{(i)}\delta_{\tilde{X}_{0:t}^{(i)}}(\mathbf{x}_{0:t})$, where $\tilde{X}_{0:t}^{(i)}=\{X_{0:t-1}^{(i)},\tilde{X}_t^{(i)}\}_{i=1}^N$;
\IF{ $\hat{S}_{eff}=\left(\sum_{i=1}^N(w_t^{(i)})^2\right)^{-1} < \frac{N}{2}$} 
\STATE $\left(\{\tilde{X}_t^{(i)}\}_{i=1}^N\right)\xrightarrow{cluster} \{C_1,...,C_{k}\}$; 
\FOR{each cluster $C_j$}
	\STATE $v_t^{(k)} = \sum_{\tilde{X}_t^{(i)}\in C_j}w_t^{(i)}$; 
	\STATE Resample ${X'}_{0:t}^{(m)}\sim \sum_{i:\tilde{X}_t^{(i)}\in C_j}\frac{w_t^{(i)}}{v_t^{(k)}}\delta_{\tilde{X}_{0:t}^{(i)}}(\mathbf{x}_{0:t})$, $m=1,...,|C_j|$
\ENDFOR
\STATE $w_t^{(i)}\leftarrow \frac{v_t^{(j)}}{|C_j|/N}\mathbb{I}({X'}_t^{(i)}\in C_j)$;
\STATE Return $\hat{p}(\mathbf{x}_{0:t}|\mathbf{y}_{1:t})=\sum_{i=1}^Nw_t^{(i)}\delta_{{X'}_{0:t}^{(i)}}\{\mathbf{x}_{0:t}\}$.
\ENDIF
\end{algorithmic}
\end{algorithm}

\begin{algorithm}[h]
\label{alg:CBF}
\caption{Clustering based Bootstrap Filter (CBF)}
\begin{algorithmic}[1]
\STATE $t=0$, Sample $X_0^{(i)} \sim p(\mathbf{x}_0)$, $i=1,\ldots,N$.
\WHILE{$t\geq1$}
\FOR{each $i$}
\STATE Sample $\tilde{X}_t^{(i)} \sim \pi(\mathbf{x}_t|\mathbf{x}_{t-1}=X_{t-1}^{(i)})$;
\ENDFOR
\STATE $\left(\{w_t^{(i)}\}_{i=1}^N,\{X_{0:t}^{(i)}\}_{i=1}^N,\hat{p}(\mathbf{x}_{0:t}|\mathbf{y}_{1:t})\right)=$ \text{WeightAdjust}$\left(\{X_{0:t}^{(i)},\tilde{X}_t^{(i)}\}_{i=1}^N\right) $;
\STATE $t \leftarrow t+1$;
\ENDWHILE
\end{algorithmic}
\end{algorithm}

Therefore, we combine the independent sub-sampling results together with a weight adjust step by rescaling. In fact, we only need to assign the cluster-specific weights $\frac{v_t^{(j)}}{|C_j|/N}$. That is to say, if ${X'}_t^{(i)} \in C_j$, then we assign $w_t^{(i)}=\frac{v_t^{(j)}}{|C_j|/N}$ instead of $\frac{1}{N}$, thus leading a new posterior estimate (\ref{eq:new}). This step will guarantee Proposition \ref{prop:expectation} holds. Algorithm \ref{alg:waa} shows one complete iteration of SIR.  
\begin{align}\label{eq:new}
\hat{p}(\mathbf{x}_{0:t}|\mathbf{y}_{1:t})=\sum_{i=1}^Nw_t^{(i)}\delta_{{X'}_{0:t}^{(i)}}(\mathbf{x}_{0:t})
\end{align}
\begin{proposition}\label{prop:expectation}
The estimated posteriors in Line 5 and 13 of Algorithm \ref{alg:waa} have the same expectation.
\end{proposition}

A parallel thought is applied to a bunch of SMCs on each equal size sub-dataset \cite{casarin2013parallel}, and the final combination is a simple average. In contrast, we use the entire dataset for SMC but parallelize the sub-resampling within each cluster. Since the size of clusters is not equal, we adopt a weighted average. The modification is equivalent to sampling from the following distribution
\begin{align}\label{eq:csmc}
\hat{p}(\mathbf{x}_{0:t}|\mathbf{y}_{1:t})=\sum_{j=1}^{k}\frac{|C_j|}{N}\sum_{i:X_{0:t}^{(i)}\in C_j}\frac{w_t^{(i)}}{v_t^{(j)}}\delta_{\tilde{X}_{0:t}^{(i)}}(\mathbf{x}_{0:t})
\end{align}
For clusters with $\frac{|C_j|}{N}>v_t^{(j)}$, this strategy will work to relieve the sample impoverishment since more particles will be sampled than it should be. Denote two categorical distributions with parameter sets $\left\{\frac{|C_j|}{N}\right\}_{j=1}^k$ and $\{v_t^{(j)}\}_{j=1}^k$, we can define their Kullback-Leibler (KL) divergence:
\begin{align}\label{eq:KL}
D_{KL}\left(\left\{v_t^{(j)}\right\}\left\|\left\{\frac{|C_j|}{N}\right\}\right.\right)=\sum_{j=1}^k v_t^{(j)}\log\frac{v_t^{(j)}}{|C_j|/N}
\end{align}
Even when Proposition \ref{prop:expectation} holds, the change of variance is difficult to obtain. But we use the KL divergence as another difference measurement, i.e. Proposition \ref{prop:KL}. Thus Equation (\ref{eq:KL}) exactly measures how the actual sampled posterior differs from the original posterior. Larger $D_{KL}$ means the interpretation of the geometric structure by SMC differs more from the clustering method.  If $D_{KL}$ happens to be 0, the sub-resampling reduces to standard resampling.  

\begin{proposition}\label{prop:KL}
The KL divergence between the actual sampled posterior in Equation (\ref{eq:csmc}) and the regular SMC estimated posterior in Line 5 of Algorithm \ref{alg:waa} equals $D_{KL}\left(\left\{v_t^{(j)}\right\}\left\|\left\{\frac{|C_j|}{N}\right\}\right.\right)$.
\end{proposition}

Alg. \ref{alg:CBF} wraps Alg. \ref{alg:waa} up with an outer loop, thus forming the clustering based bootstrap filtering. In fact, it is unnecessary to specify a clustering method, because the accuracy of clustering does not matter if the purpose is to increase the sampling odds of small weights particles. $k$-means, as a simple and fast clustering algorithm, is appropriate to implement in CBF. Even for high dimensional data, $k$-means after a PCA preprocessing step makes sense, since \cite{drineas2004clustering} proved a theorem implying that $k$-means clustering on the top $k$-dimensional subspace associated with PCA is at most 2 times worse than the optima of the original space. This is the reason we suggest $k$-means as a generic method. In addition, the number of clusters $k$ also plays an unimportant role, though should be relatively large, since more discovered sub-clusters are allowed. Additionally, a high dimensional extension from Bayesian perspective is discussed in Appendix.

\section{Experiments}
\label{sec:exp}

In this section, we first investigate how SSMC can contribute to the initialization of $k$-means via a simulation study.  

\subsection{Simulation Study} 
\begin{figure*}[ht]
\centering
\subfigure[$t=0$]{
\includegraphics[width=30mm,clip,trim=50 15 50 15mm]{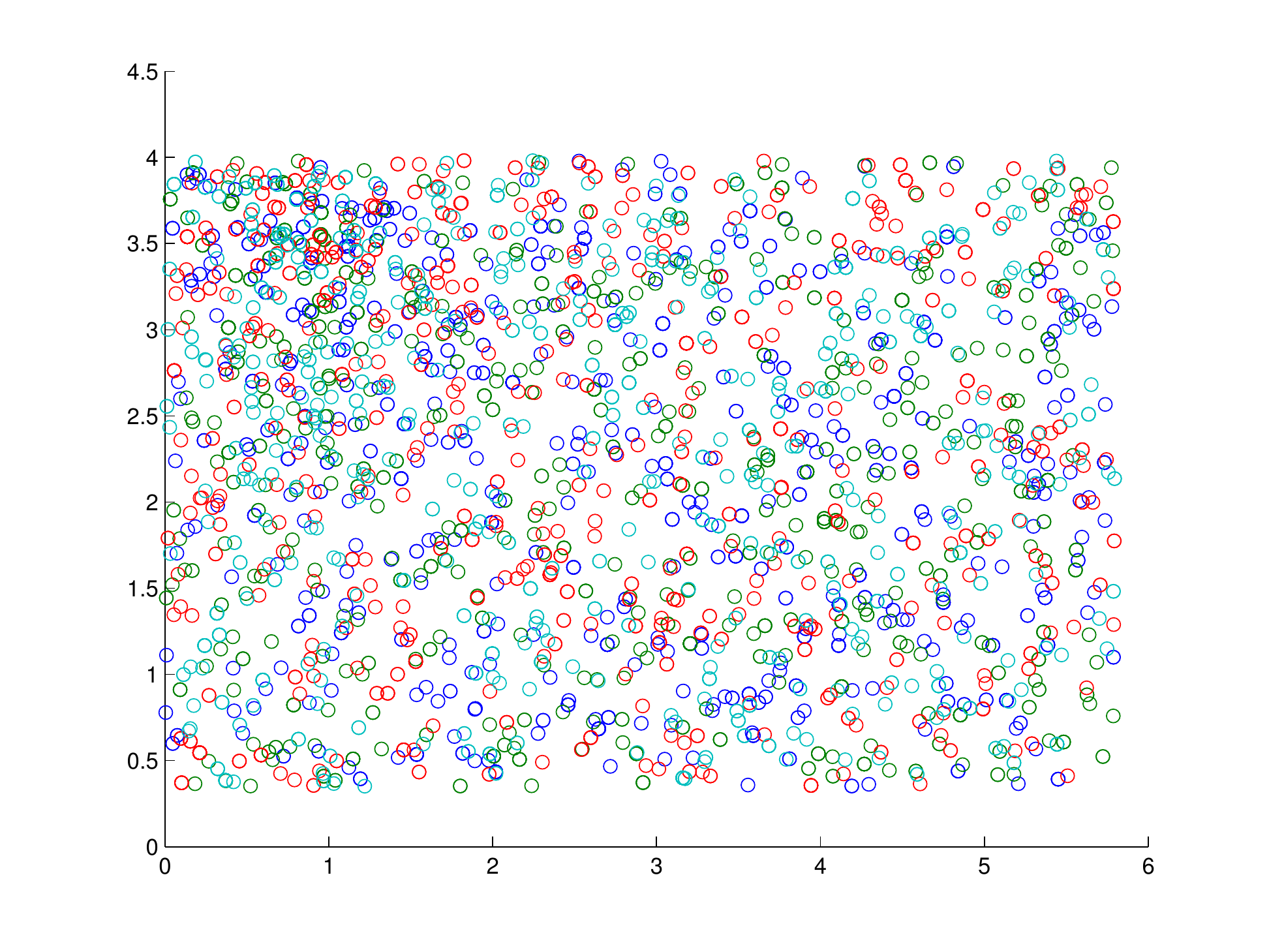}}
\subfigure[$t=4$]{
\includegraphics[width=30mm,clip,trim=50 15 50 15mm]{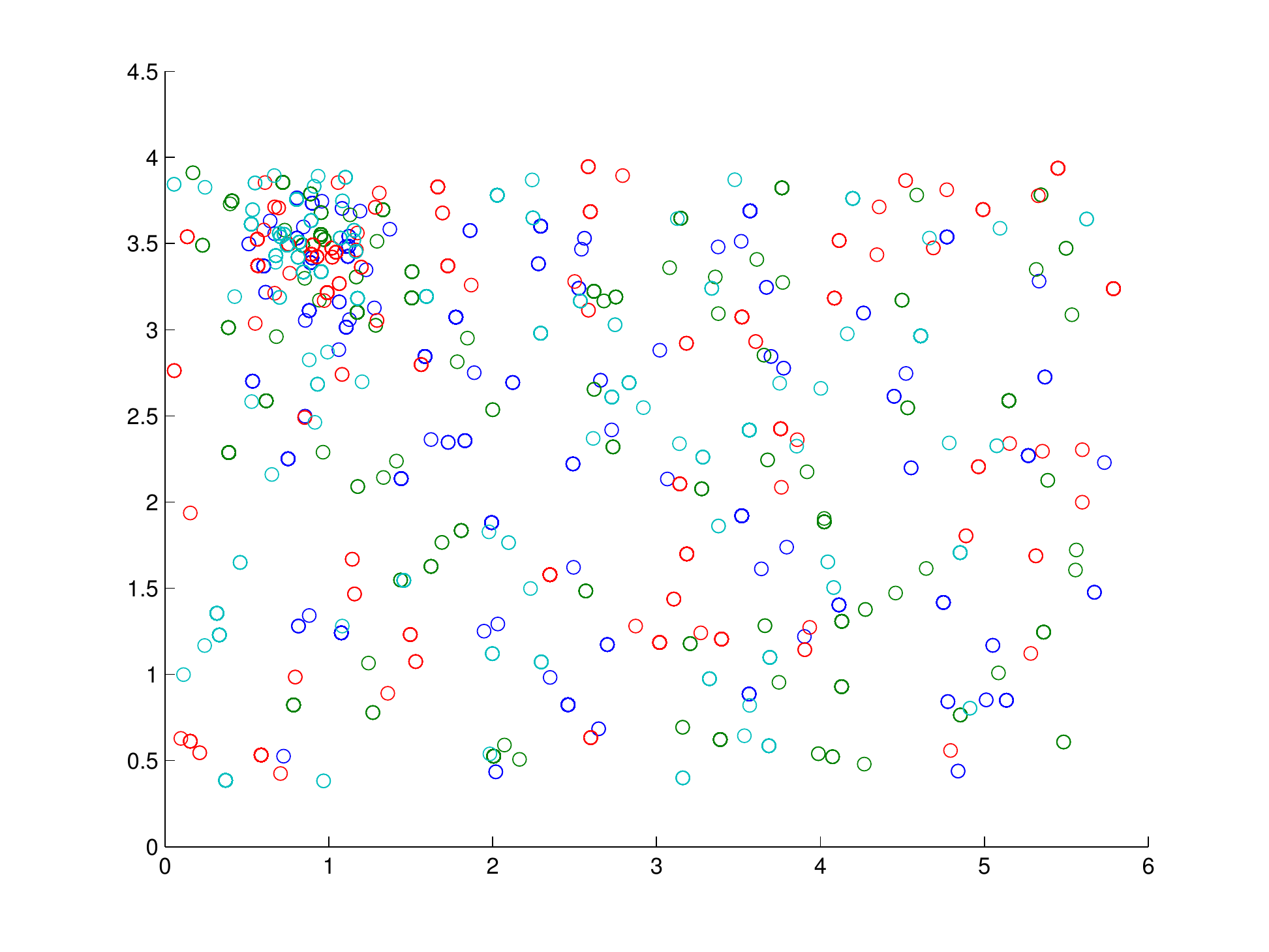}}
\subfigure[$t=8$]{
\includegraphics[width=30mm,clip,trim=50 15 50 15mm]{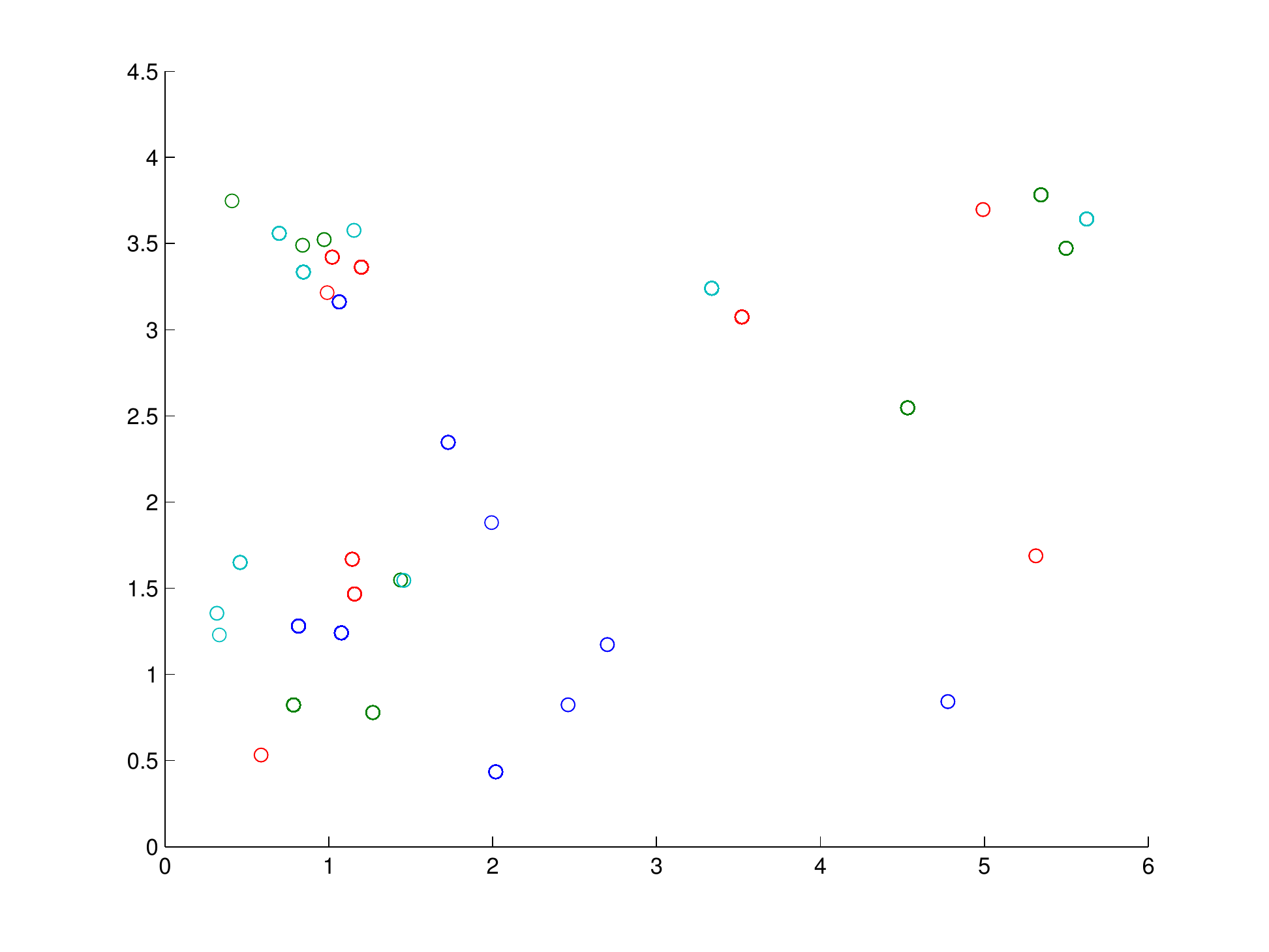}}
\subfigure[$t=18$]{
\includegraphics[width=30mm,clip,trim=50 15 50 15mm]{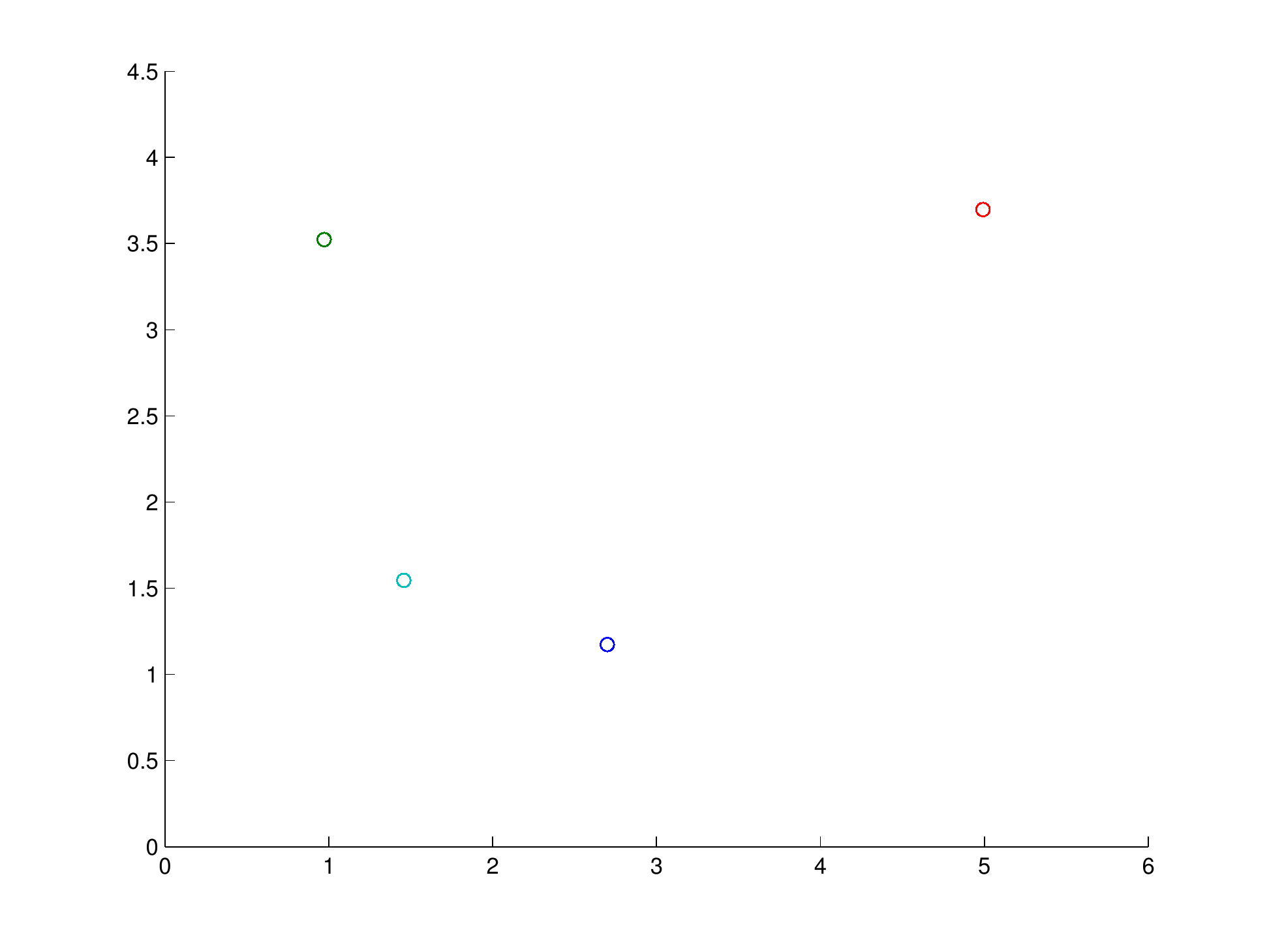}}
\subfigure[Result Clusters]{
\includegraphics[width=30mm,clip,trim=50 15 50 15mm]{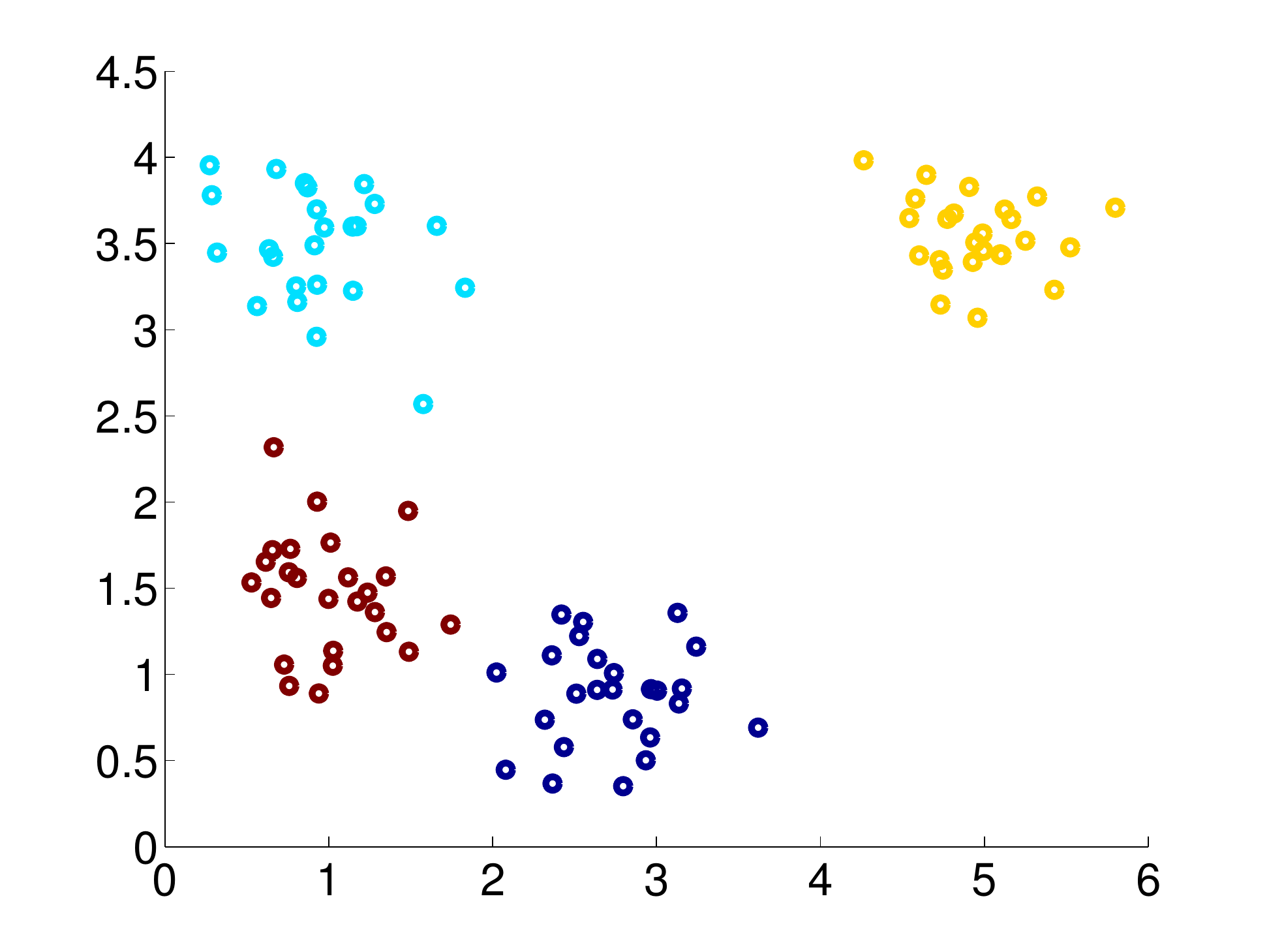}}
\centering
\caption{(a) - (d) show how particles change over iterations of SSMC. (a) initial particles; (b) and (c) illustrate how particles degenerate to potential clusters; (d) final collapsed particle; (e) is the resulting clusters of $k-$means initialized by (d).}
\label{fig:ssmc}
\end{figure*}

We generate four equal clusters of 100 points from a mixture of 2-d Gaussian distributions $\mathcal{N}(\mu_j,\sigma^2\mathbf{I}_2)$, $j=1,\cdots,4$, where $\mu_1=(0.7,3.5)$, $\mu_2=(1,1.5)$, $\mu_3=(2.7,1)$, $\mu_4=(5,3.5)$ and $\sigma^2=0.1$. Figure \ref{fig:bad}(a) is also generated from this setup. For simulation, we assume $\sigma^2$ is fixed or known to the model, which means it is not required to be estimated. This will be beneficial for visualizing the rationale of SSMC. The batch size used int the SSMC algorithm is 10, indicating one tenth of observations are randomly chosen for training in each iteration. Figure \ref{fig:ssmc}(a) - (d) show the process of SSMC, beginning with randomly initialized particles in the bounded space of the data domain and ending with unique set of parameters. Degeneracy is present with sample impoverishment over iterations, showing the collapsing to cluster means. The final obtained particle is not necessarily the exact cluster means, the purpose is to avoid the ill-posed initialization in Figure \ref{fig:bad}(b).

\begin{table}[t]
\caption{A Simulation Study}
\label{tab:simul}
\begin{center}
\begin{tabular}{lrr}
\multicolumn{1}{c}{\bf Algorithm}  &\multicolumn{1}{c}{\bf Failure Rate} &\multicolumn{1}{c}{\bf 1st Failure}\\
\hline 
$k$-means         & 26.7\% (4/15) & 3rd test\\
$k$-means++		  &   20\% (3/15) & 4th test\\
SSMC+$k$-means    &  6.7\% (1/15) & 15th test\\
\end{tabular}
\end{center}
\end{table}

As our knowledge of other mechanisms in terms of fighting degeneracy, no theoretically guaranteed quantity is appropriate to measure our approach. Thus in our simulation study, we intend to evaluate how likely the clustering is to fail (i.e. local minima shown in Figure \ref{fig:bad}(b)) will happen. Using the same generation model to synthesize data, we first run our SSMC initialization and then a standard $k$-means algorithm, and discover that the first local minima occurs until the 15th test. For comparison, we also run standard $k$-means with random initialization and $k$-means++ (default implemented in Matlab) 15 times. The results of this simulation study are shown in Table \ref{tab:simul}. It is apparent from this table that the failure probability of  SSMC initialized $k$-means is much lower than other two algorithms. It is suggested in general that our approach is an alternative initialization method for $k$-means.

\begin{figure}[tb]
\centering
\subfigure[BF]{
\includegraphics[width=40mm,clip,trim=50 15 50 15mm]{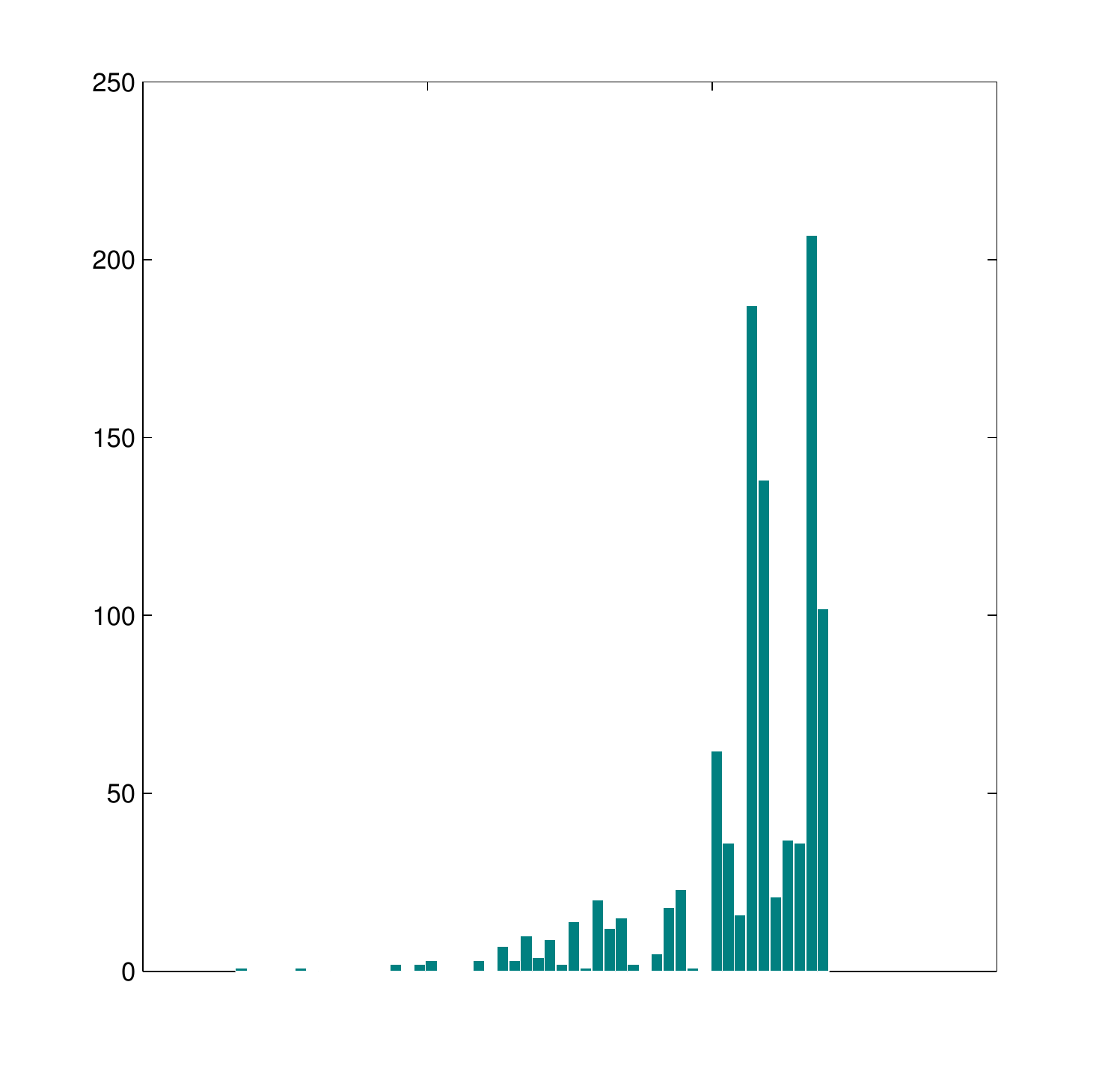}}
\subfigure[CBF]{
\includegraphics[width=40mm,clip,trim=50 15 50 15mm]{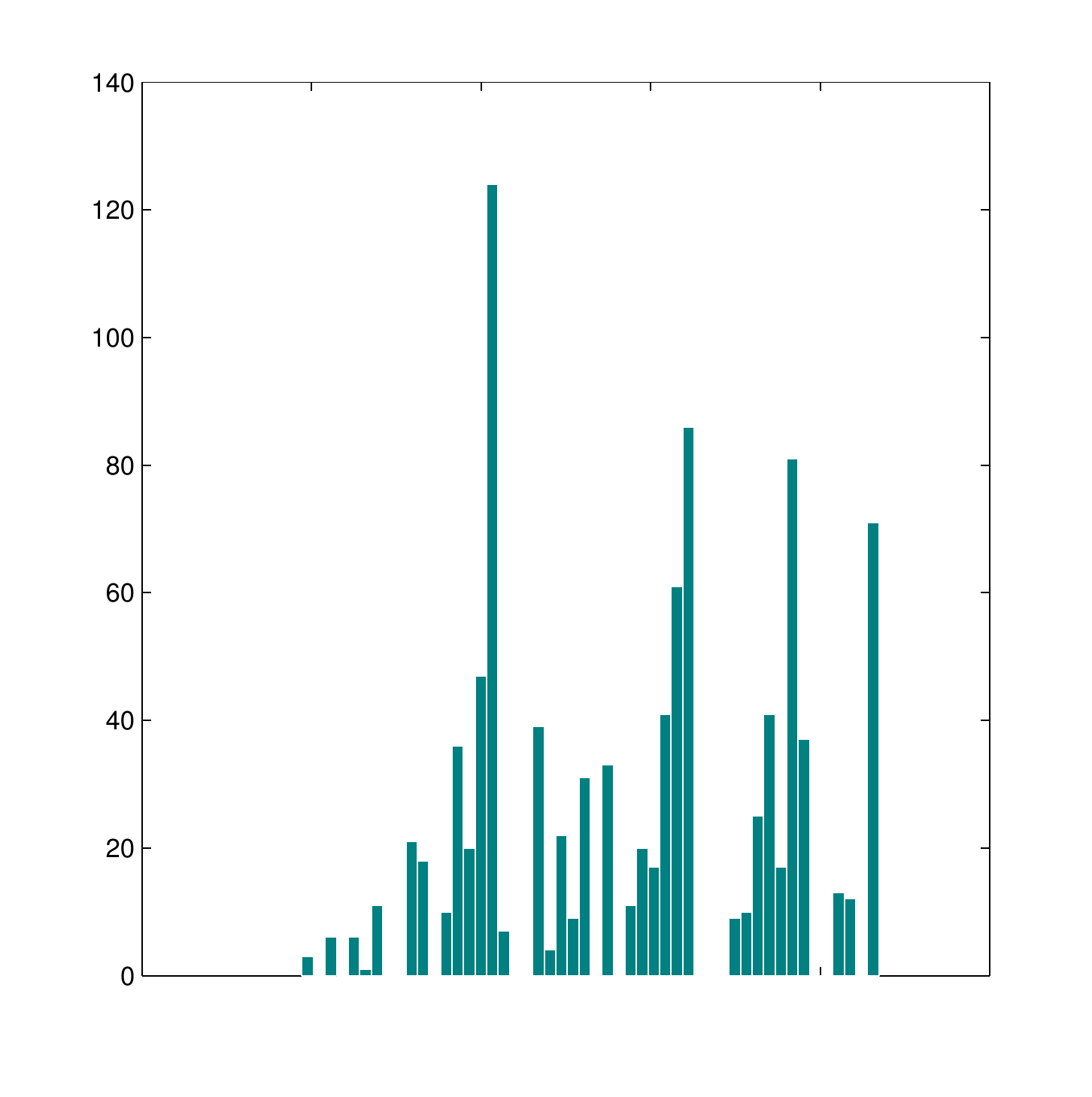}}
\caption{Particles Histogram Distribution}
\label{fig:pf}
\end{figure}

We consider another popular stochastic volatility model as described by \cite{durbin2000time}.
\begin{align*}
x_t &= \phi x_{t-1} + \sigma\epsilon_t, x_1\sim\mathcal{N}(0,\sigma^2/(1-\phi^2))\\
y_t &= \beta\exp(x_t/2)\eta_t
\end{align*}
where $\epsilon_t$ and $\eta_t$ follows a standard normal distribution. In our simulation, we set $\sigma^2=0.9$, $\phi=0.8$, $\beta=0.7$ and $T=40$. The number of particles is $N=1000$. We run standard bootstrap filtering and $k$-means clustering based filtering by setting $k=10$. Figure \ref{fig:pf} displays the particle histogram distribution at time $T$, i.e. $\hat{p}(x_T|y_{1:T})$, supporting the intuition that CBF can postpone sample impoverishment. The degeneracy problem appears for the BF algorithm estimated posterior, while CBF attains more modes. In the next example, we apply this to a multiple path tracking problem.

\subsection{MH 370 Multiple Potential Paths Tracking}

We then explore how a clustering based bootstrap filtering algorithm can be used for real Jet tracking, which is extensively used in civil and military aviation, and an automated technique that allows the remote control center to detect the trajectory of the aircraft based on received radar data is necessary. With the advent of recent aviation emergencies, jet tracking without prior knowledge of starting point is emphasized. For example, the mysterious vanishing of MH 370 was linked with almost no information about where it disappeared so that there is a completely contradictory debate on the possibility of south and north paths.

Particle filtering methods have been widely used in tracking problems. In our setting of aircraft tracking, we assume that we do \textbf{not} know where the object starts in a known map. However our radar sensor can obtain data on the height of the jet above ground and its horizontal velocity. For simplicity, we further assume the time series radar data is discrete and equal time intervals between every two successive signals, and the aircraft will not change its horizontal velocity during each time period, i.e. observed radar data $\mathbf{y}_{1:T}=\{(h_t,\mathbf{v}_t)\}_{t=1}^T$.

\begin{figure}[htb]
\centering
\subfigure{
\includegraphics[width=60mm,clip,trim=70 40 55 30mm]{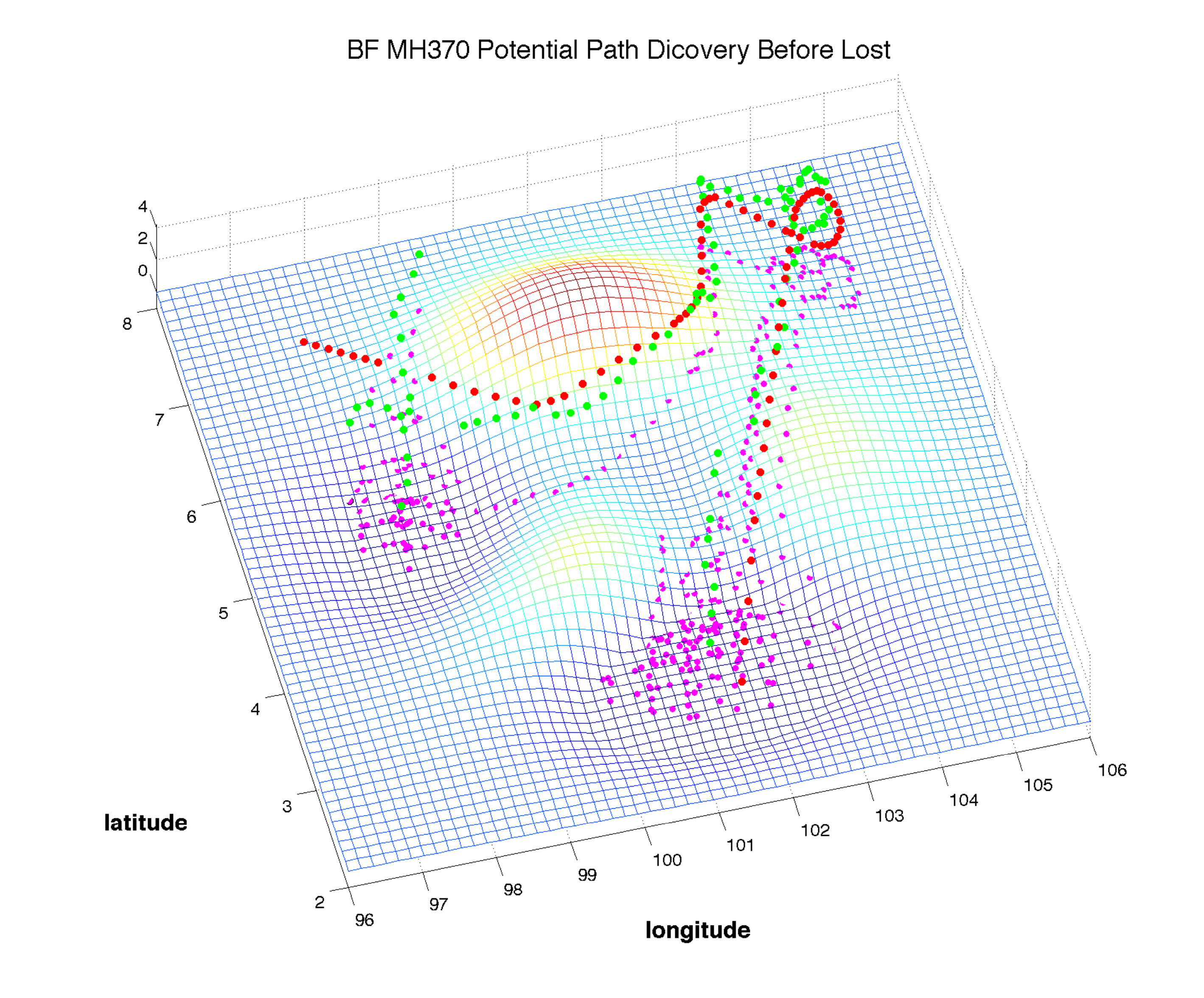}}
\subfigure{
\includegraphics[width=60mm,clip,trim=75 50 55 35mm]{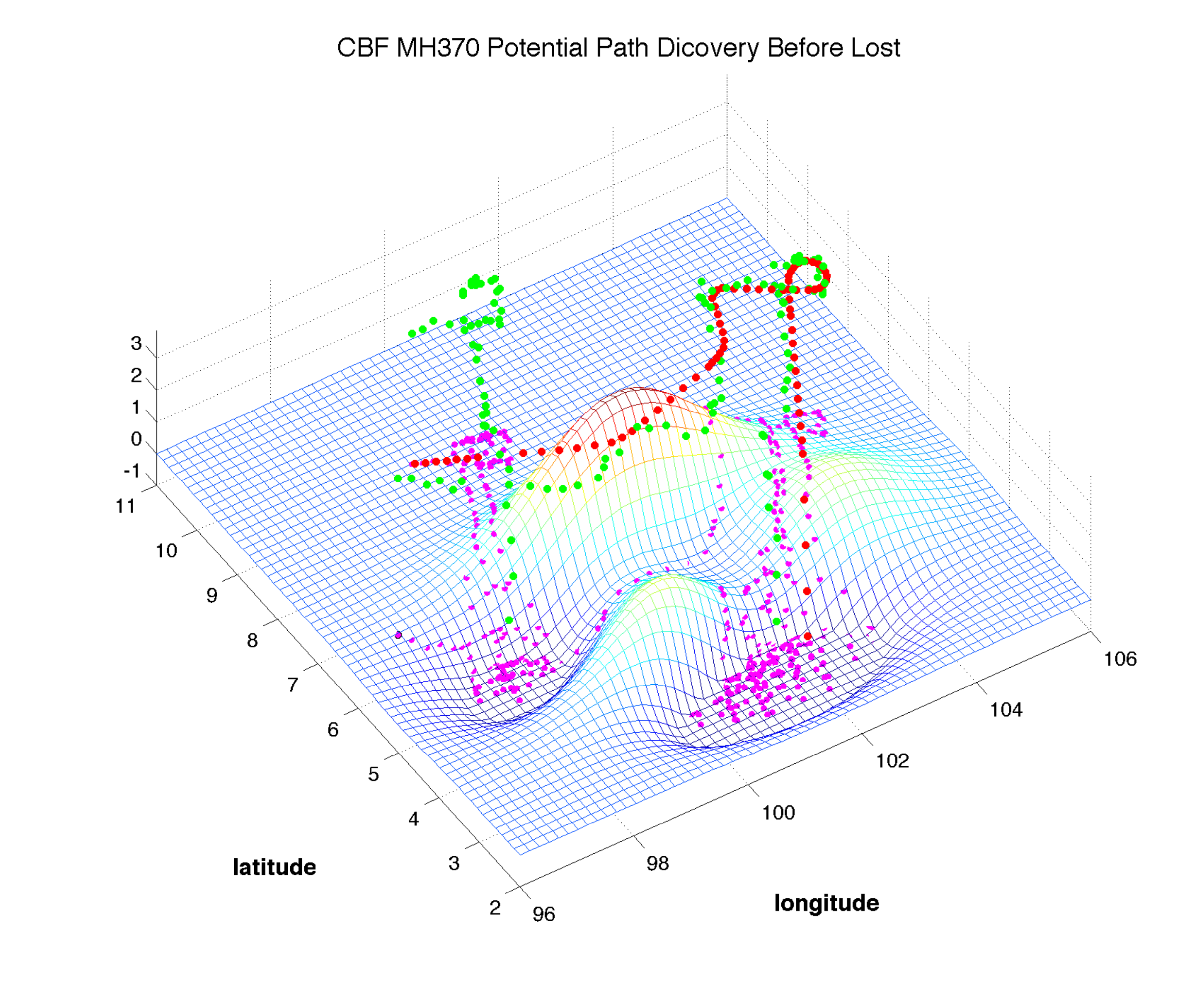}}
\centering
\caption{Red dots: True trajectory of MH 370; Purple dots: Particles in each iteration; Green dots: estimated trajectory (i.e. estimated posterior mean). CBF (right) algorithm discovers the potential paths longer than standard BF (left).}
\label{fig:mh}
\end{figure}

We assume that hidden state $\mathbf{x}_t=(x_1,x_2)_t$ is the longitude and latitude pair at time $t$. Initially, $\mathbf{x}_0^{i}\sim$ Uniform$(A)$, where $A$ is the area of the map. Let $Z(x_1,x_2)$ be the terrain function, which maps the longitude and latitude to the ground altitude. The model is:
\begin{align*}
\mathbf{x}_t &= \mathbf{x}_{t-1} + \mathbf{v}_t\Delta t + \epsilon_x, \epsilon_x\sim \mathcal{N}(\mathbf{0},\Sigma)\\
h_t &= h - Z(\mathbf{x}_t) + \epsilon_h, \epsilon_h\sim \mathcal{N}(0,\sigma_h^2)
\end{align*}
where $\epsilon_x$ and $\epsilon_h$ are radar noise. $\epsilon_x$ does exist because $\mathbf{v}_t$ is one element of radar data $\mathbf{y}_t$ that has some device noise. $h$ is cruising altitude above sea level. Then we could easily use the likelihood of $h_t$ to update the weights. The terrain is smoothed for simplicity as shown in Fig.\ref{fig:mh}. 

We marked the real flight trajectory (longitude and latitude) of MH370 captured by radar (before it went missing) using Google Earth software. Based on the trajectory, we generated discrete sequential signals $\{\mathbf{y}_t\}_{t=1}^T$. $\mathbf{v}_t$ is computed by the displacement of two trajectory points and $\Delta t$. According to the result\footnote{BF: \url{https://youtu.be/bFx2DeengXY}; CBF: \url{https://youtu.be/pgC3f65y_ZE}.} illustrated in Fig.\ref{fig:mh}, the red dots notating location is the ground truth (real path of flight). The purple dots, the projection onto the ground of the randomly sampled locations of potential paths, represent the next-step predictive guesses based on current radar sensor data, which will be modified once new data streams in. The green dots are the final estimated paths based on all radar record. Since the whole tracking process should be dynamic with streamed data, the figures merely show the final snapshot when all of the data is received. To be more specific, people can imagine red and green dots  appear one by one at each time stamp, while the purple dots appear a few at a time during each time interval.
 
It is noticeable that the left path of the BF algorithm is shorter than the one of CBF, meaning that the traditional method loses one possible path earlier. It makes more sense to discover more possible paths in detecting a lost plane or enemy plane in a military application. Under such circumstances, it is too risky to miss detecting one possible path. CBF is a more conservative but safer compromise in false positive discovery. Still taking the example of MH 370, the obtainable information is merely the handshake data per hour (signal sending and receiving) between artificial satellite and the flight, including extra information like Doppler effects to deduce the radial velocity of aircraft, but no vanishing location. This setting is exactly like the trajectory tracking without a starting point, thus allowing the same algorithm to be used. 

\section{Conclusion}
\label{sec:con}

In this paper, we investigate how to bridge clustering methods and sequential Monte Carlo, and how they may benefit each other; stochastic SMC can contribute to a robust initialization for $k$-means algorithm, while clustering based bootstrap filtering arises the hope of fixing degeneracy problems in SMC. The experimental test shows both these approaches outperform the vanilla algorithms, showing that there are improvements present in these techniques. In the future further experimental results will elucidate the benefits of this method. Another future direction might explore implementing a parallel method to speed up the algorithm, thus making it more applicable in the real world. 

\bibliography{template.bib}
\bibliographystyle{plain}

\newpage

\section{Appendix}

The subspace in (Drineas, 2004) is derived from the PCA, causing fixed sub-dimensionality for all clusters. However, this geometric structure is not necessarily true, especially to high dimensional data. Each cluster may be associated with different relevant dimensions, meaning they exist in different subspaces. Take one extreme example, for a $D$ dimensional dataset, there are $D-1$ subspace clusters generating from a 2 dimensional Gaussian distribution in dimension $d$ and $d+1$, $d=1,...,D-1$, while other irrelevant dimensions within each cluster are uniform distributed noise. In this example, $k$-means will be likely to fail since the Euclidean measure fails due to curse of dimensionality. PCA will also fail due to nonexistence of dominant subspace. In order to grasp the true clusters as accurate as possible, we briefly introduce how to leverage more general subspace clustering method to construct the similar CBF Algorithm.  

If dimensionality $d$ is not to small compared with the number of particles $N$, finding a partition of dimensions for each cluster makes more sense to represent the geometric structure. This intuition is quite similar to group Lasso (Yuan, 2006) but our setting is under unsupervised learning. For simplicity, the subspace cluster model proposed in this paper shall only consider non-overlapping sub-dimensionality, although more subtle analysis can be adapted to handle moderately overlapping groups. (Hoff, 2006) proposed a non-parametric subspace clustering algorithm under the assumption of local Gaussian distributed data (similar to the extreme case above). Generally, we ignore Gaussian constrain for the sub-modes (or clusters) finding, because the distribution around the mode can be treated as Gaussian by Laplace Approximation.

\paragraph{Subspace Clustering} We restrict the idea of (Hoff, 2006) on fixed $k$ clustering. Consider the linear decomposition model as follows,
\begin{align}\label{eq:submodel}
X_t^{(i)}=\mu+\mathbf{r}^{z_i}\times\mathbf{b}^{z_i}+\epsilon_i
\end{align}
where $\mu\in\mathbb{R}^d$ describes the global mean of the whole space, $\mathbf{r}^{z_i}\in\{0,1\}^d$ indicates \textit{relevant dimension} of cluster $z_i\in[k]$, $\mathbf{b}^{z_i}\in\mathbb{R}^d$ represents the mean shifts away from $\mu$ on cluster-specific relevant dimension, and $\epsilon_i$ are i.i.d $d$ dimensional zero mean Gaussian noise. It is noted that $\times$ means Hadamard (element-wise) multiplication. Thus $\mu_j=\mu+\mathbf{r}^j\times\mathbf{b}^j$ represents the mean of $j$th cluster. This is a more general Gaussian Mixture model; it will reduce to standard GMM if $\mathbf{r}=\mathbf{1}$; it will achieve a PCA-like dimensionality reduction if fixing $\mathbf{r}$. Compared with the nonparametric version in (Hoff, 2006), this setting can both simplify the model and speed up the computation. The bayesian formulation can achieve the cluster assignment and cluster-specific relevant dimensions simultaneously (see later derivation). 

\paragraph{Posterior Construction} An observation in the subspace clustering is that we obtain the results with information of relevant dimensions $\mathbf{r}_t^j$ in each cluster at the same time. We can take the advantage of this information to design more complicated $\hat{p}$. For each $X_t^{(i)}$, there exists $z_i\in[k]$ such that $X_t^{(i)}\in C_{z_i}$. In other words, $\mathbf{r}_t^{z_i}$ represents the relevant dimension of $X_t^{(i)}$. The projection of $X_t^{(i)}$ into $\mathbf{r}_t^{z_i}$ is denoted as $X_{t,\mathbf{r}_t^{j_i}}^{(i)}$. During the sub-resampling step, one possible adjustment for posterior in Algorithm \ref{alg:waa} follows.
\begin{align}\label{eq:highsmc}
&\hat{p}_{C_j}(\mathbf{x}_{0:t}|\mathbf{y}_{1:t})=\\
&\sum_{\tilde{X}_t^{(i)}\in C_k}\left(\frac{w_t^{(i)}}{v_t^{(k)}}\delta_{\tilde{X}_{0:t,\mathbf{r}_t^{j_i}}^{(i)}}(\mathbf{x}_{0:t,\mathbf{r}_t^{z_i}}),w_t^{(i)}\delta_{\tilde{X}_{0:t,\overline{\mathbf{r}_t^{j_i}}}^{(i)}}(\mathbf{x}_{0:t,\overline{\mathbf{r}_t^{z_i}}})\right) \nonumber
\end{align}
where $\overline{\mathbf{r}_t^{z_i}}$ represents the irrelevant dimensions of $C_{j_i}$.

This weight updating strategy allows the weights diversity exists in different dimensions for each particle. It makes more sense because submodes do exist in the relevant dimensions of high dimensionality. However, we need to be careful during sub-resampling, since new particles may create due to sampling separately across different dimensions, which has the same effect of constructing Markov kernel (Gilks, 2001). Fortunately, new generated particles can contribute to relieve the degeneracy problem, but the consistence property of this new posterior is still unknown. 

\subsection{Derivation of Bayesian Subspace Clustering with $k$ Memberships}

We restrict the idea of (Hoff, 2006) on fixed $k$ clusters setting. But for simplicity, we omit the time index $t$ here, denote the latent variable $z_i$ indicating the instance membership. Consider the linear decomposition model again,
\begin{align}\label{eq:submodel}
\mathbf{x}_i=\mu+\mathbf{r}^{z_i}\times\mathbf{b}^{z_i}+\epsilon_i
\end{align}
where $\mu\in\mathbb{R}^d$ is the global mean of the whole space, $\mathbf{r}^{z_i}\in\{0,1\}^d$ is \textit{relevant dimension} of cluster $z_i\in[k]$, $\mathbf{b}^{z_i}\in\mathbb{R}^d$ describes the mean shifts away from $\mu$ on cluster-specific relevant dimension, and $\epsilon_i$ are i.i.d $d$ dimensional zero mean Gaussian noise. It is noted that $\times$ means element wise multiplication. Thus $\mu_j=\mu+\mathbf{r}^j\times\mathbf{b}^j$ represents the mean of $j$th cluster. This is a more general Gaussian Mixture model, and it will reduce to standard GMM if $\mathbf{r}$ always equals to all 1s vector. 

\subsection{Generative Model}
\begin{enumerate}
\item For $j=1,...,k$,
\begin{align*}
\mathbf{r}^j=(\mathbf{r}_m^j)_{m=1}^d&\sim \prod_{m=1}^d \text{Ber}(\theta_m)\\
\mathbf{b}^j=(\mathbf{b}_m^j)_{m=1}^d&\sim \prod_{m=1}^d \mathcal{N}(0,\tau_m^2)
\end{align*}
\item For $i=1,...,N$
\begin{align*}
z_i&\sim\text{Mul}(\pi),\sum_{j=1}^k\pi_j=1\\
\mathbf{x}_i&\sim\mathcal{N}(\mu+\mathbf{r}^{z_i}\times\mathbf{b}^{z_i},\Sigma)
\end{align*}
where $\mu$ is d dimensional vector $(\mu_m)_{m=1}^d$, and $\Sigma$ usually takes the diagonal matrix with element $\{\sigma_1^2,...,\sigma_d^2\}$.
\end{enumerate}
 
This model seems quite flexible in various dimensions; however, we expect a less overfitting model, thus suggesting $\theta$ is not indexed by $m$ and $\tau_m^2=\eta\sigma_m^2$. Therefore, we can formulate our bayesian generative model as follows.
\begin{enumerate}
\item $\theta\sim\text{Beta}(a_\theta,b_\theta)$
\item $\eta\sim\text{IG}(a_\eta,b_\eta)$
\item $\pi\sim\text{Dir}(\alpha)$
\item For $m=1,...,d$,
\begin{align*}
\mu_m&\sim\mathcal{N}(m,v) \\
\sigma_m^2&\sim\text{IG}(a_\sigma,b_\sigma)
\end{align*}
\item For $j=1,...,k$,
\begin{align*}
\forall m\in[d],\mathbf{r}_m^j&\sim  \text{Ber}(\theta)\\
\forall m\in[d],\mathbf{b}_m^j&\sim  \mathcal{N}(0,\eta\sigma_m^2)
\end{align*}
\item For $i=1,...,N$
\begin{align*}
z_i&\sim\text{Mul}(\pi),\sum_{j=1}^k\pi_j=1\\
\mathbf{x}_i&\sim\mathcal{N}(\mu+\mathbf{r}^{z_i}\times\mathbf{b}^{z_i},\text{diag}\{\sigma_1^2,...,\sigma_d^2\})
\end{align*}
\end{enumerate}
where IG means inverse gamma distribution.

\subsection{Parameter Estimation by Gibbs Sampler}
We can compute the conditional likelihood as follows.
\begin{align*}
P(X|Z)&=\prod_{j=1}^k\prod_{m=1}^d\left[\theta\mathcal{N}\left(x_{i,m:z_i=j}|\mu_m\mathbf{1}_{n_j},\sigma_m^2(\mathbf{I}_{n_j}+\eta\mathbf{1}_{n_j}\mathbf{1}_{n_j}')\right)\right.\\
&\left.+(1-\theta)\mathcal{N}\left(x_{i,m:z_i=j}|\mu_m\mathbf{1}_{n_j},\sigma_m^2\mathbf{I}_{n_j}\right)\right]\\
&=\left[\prod_{j=1}^k\prod_{m=1}^d((1-\theta)+\theta\lambda_m^j)\right]\left[\prod_{i=1}^N\prod_{m=1}^d\mathcal{N}(x_{i,m}|\mu_m,\sigma_m^2)\right]
\end{align*}
where $\lambda_m^j=\frac{\mathcal{N}\left(x_{i,m:z_i=j}|\mu_m\mathbf{1}_{n_j},\sigma_m^2(\mathbf{I}_{n_j}+\eta\mathbf{1}_{n_j}\mathbf{1}_{n_j}')\right)}{\mathcal{N}\left(x_{i,m:z_i=j}|\mu_m\mathbf{1}_{n_j},\sigma_m^2\mathbf{I}_{n_j}\right)}=\sqrt{\frac{1}{1+\eta n_j}}\exp\left\{\frac{\eta n_j}{\eta n_j+1}\sum_{i:z_i=j}\frac{(x_{i,m}-\mu_m)^2}{2\sigma_m^2}\right\}$. The second equation may need the fact $\text{Det}(a\mathbf{I}+b\mathbf{1}\mathbf{1}')=a^n+na^{n-1}b$.

Then we can design the sampler for $z$ by integrating out $\pi$.
\begin{align*}
p(z_i=j|Z_{-i},X)&\propto(n_{-i,j}+\alpha)\prod_{m=1}^d\frac{(1-\theta)+\theta\lambda_{+i,m}^j}{(1-\theta)+\theta\lambda_{-i,m}^j}
\end{align*}
where the subscript $+i,-i$ means to include or exclude the instance $\mathbf{x}_i$ in cluster $j$.

Furthermore,
\begin{align*}
\sigma_m^2&\sim\text{IG}\left(a_\sigma+\frac{N+k}{2},\right.\\
&\left. b_\sigma+\frac{1}{2}\left[\sum_{i=1}^N(\mu_m-\epsilon_{i,m})^2+\sum_{j=1}^k\frac{\mathbf{b}_m^j}{\eta}\right]\right)\\
\mu_m&\sim\mathcal{N}\left(\hat{m},\hat{v}\right)
\end{align*}
where the residual $\epsilon_{i,n}=x_{i,m}-\mathbf{r}_m^{z_i}\times\mathbf{b}_m^{z_i}$, and $\hat{m}=\hat{v}\left(\sum_{i=1}^N\frac{\epsilon_{i,m}}{\sigma_m^2}+\frac{m}{v}\right),\hat{v}=(N\sigma_m^{-2}+v^{-1})^{-1}$.

For hyperparameters,
\begin{align*}
\theta&\sim\text{Beta}\left(a_\theta+\sum_{j=1}^k\sum_{m=1}^d\mathbf{r}_m^j,b_\theta+\sum_{j=1}^k\sum_{m=1}^d(1-\mathbf{r}_m^j)\right)\\
\eta&\sim\text{IG}\left(a_\eta+\frac{Nk}{2},b_\eta+\frac{1}{2}\sum_{j=1}^k\sum_{m=1}^d\frac{\mathbf{b}_m^j}{\sigma_m^2}\right)
\end{align*}
This finish the Gibbs sampling process. This $k$-means subspace clustering method is a straightforward generalization of Gaussian Mixture models (GMMs), like (Hoff, 2006) extended the work of (Rasmussen, 1999). 

In addition, the byproduct of this method is
\begin{enumerate}
\item Setting $\mathbf{r}=\mathbf{1}$, this model will reduce to standard GMMs.
\item Setting fixed $\mathbf{r}$, this model will reduce to GMMs on a fixed subspace, equivalent to operate a principal component analysis (PCA) to obtain a lower dimension space.
\item The advantage of 2 is the algorithm can select the dimensionality of subspace automatically. 
\end{enumerate}

\end{document}